\documentclass[11pt,a4paper]{article}
\usepackage[hyperref]{emnlp2020}
\usepackage{times}
\usepackage{latexsym}

\usepackage{microtype}

\usepackage{multirow}
\usepackage{mathrsfs}
\usepackage{amsfonts,amssymb}
\usepackage{amsmath}
\usepackage{booktabs}
\usepackage{graphicx}
\usepackage{subfigure}
\usepackage{diagbox}
\aclfinalcopy % Uncomment this line for the final submission
\def\aclpaperid{1175} %  Enter the acl Paper ID here

\title{Filtering before Iteratively Referring for Knowledge-Grounded \\ Response Selection in Retrieval-Based Chatbots}

\author{Jia-Chen Gu$^1$, Zhen-Hua Ling$^1$\thanks{\hspace{1.5mm}Corresponding author.}, Quan Liu$^{1,2}$, Zhigang Chen$^2$, Xiaodan Zhu$^3$ \\
  $^1$National Engineering Laboratory for Speech and Language Information Processing, \\
      University of Science and Technology of China, Hefei, China \\
  $^2$State Key Laboratory of Cognitive Intelligence, iFLYTEK Research, Hefei, China \\
  $^3$ECE \& Ingenuity Labs, Queen's University, Kingston, Canada \\
{\tt gujc@mail.ustc.edu.cn}, {\tt \{zhling, quanliu\}@ustc.edu.cn}, \\ {\tt zgchen@iflytek.com}, {\tt xiaodan.zhu@queensu.ca}
}

\date{}

\begin{document}
\maketitle
\begin{abstract}
  The challenges of building knowledge-grounded retrieval-based chatbots lie in how to ground a conversation on its background knowledge and how to match response candidates with both context and knowledge simultaneously.
  This paper proposes a method named \textbf{F}iltering before \textbf{I}teratively \textbf{RE}ferring (\textbf{FIRE}) for this task.
  In this method, a context filter and a knowledge filter are first built, which derive knowledge-aware context representations and context-aware knowledge representations respectively by global and bidirectional attention.
  Besides, the entries irrelevant to the conversation are discarded by the knowledge filter.
  After that, iteratively referring is performed between context and response representations as well as between knowledge and response representations, in order to collect deep matching features for scoring response candidates.
  Experimental results show that FIRE outperforms previous methods by margins larger than 2.8\% and 4.1\% on the PERSONA-CHAT dataset with original and revised personas respectively, and margins larger than 3.1\% on the CMU\_DoG dataset in terms of top-1 accuracy.
  We also show that FIRE is more interpretable by visualizing the knowledge grounding process.
\end{abstract}

\section{Introduction}

  Building a conversational agent with intelligence has received significant attention with the emergence of personal assistants such as Apple Siri, Google Now and Microsoft Cortana.
  One approach is to building retrieval-based chatbots, which aims to select a potential response from a set of candidates given the conversation context \cite{DBLP:conf/sigdial/LowePSP15,DBLP:conf/acl/WuWXZL17,DBLP:conf/acl/WuLCZDYZL18,DBLP:conf/cikm/GuLL19,DBLP:conf/acl/TaoWXHZY19,DBLP:conf/cikm/GuLLLSWZ20}.

  \begin{figure}[!hbt]
    \centering
    \includegraphics[width=7.5cm]{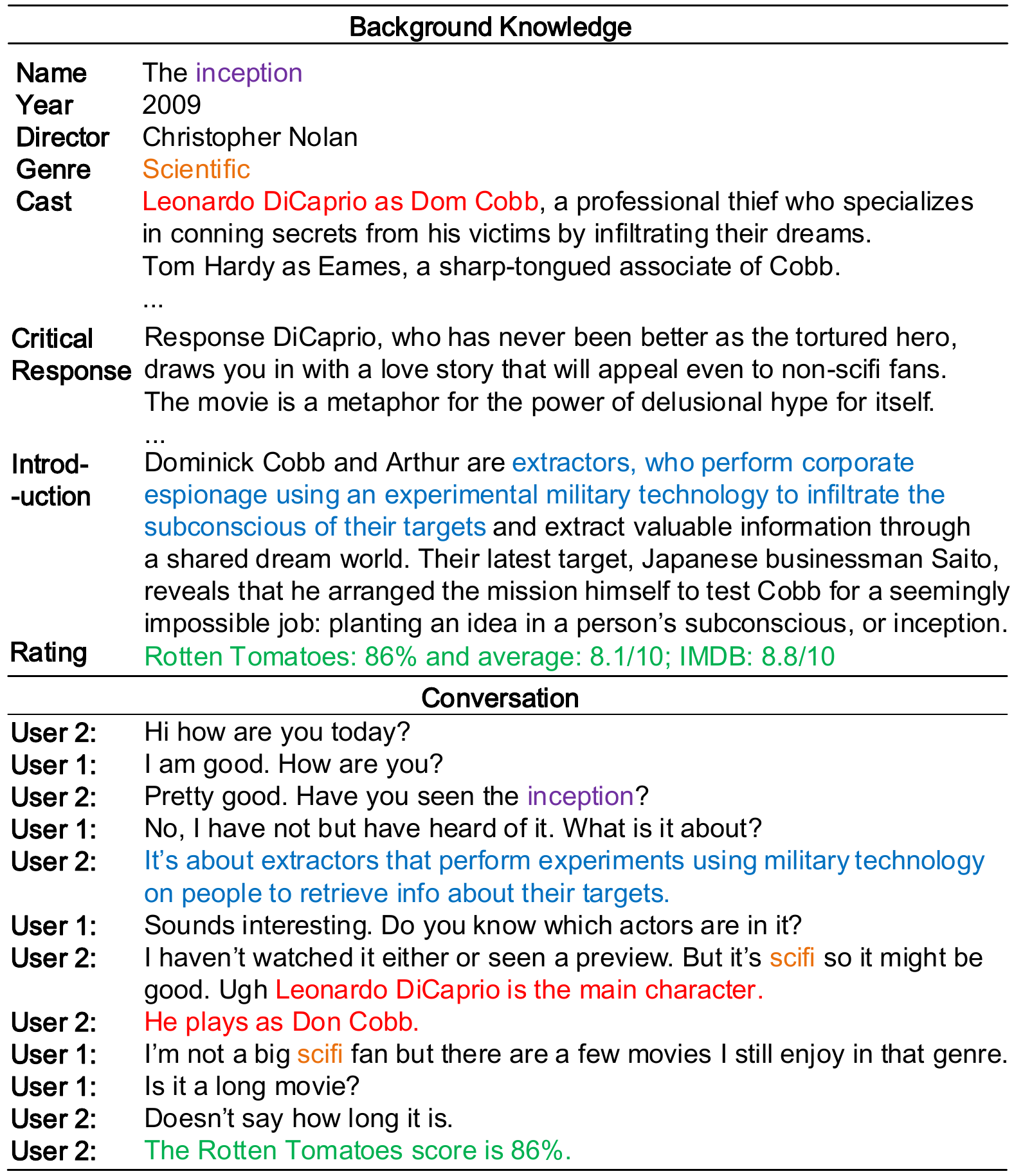}
    \caption{An example from CMU\_DoG dataset \cite{DBLP:conf/emnlp/ZhouPB18}.
             Words in the same color are related.}
    \label{fig1}
  \end{figure}

  However, real human conversations are often grounded on external knowledge.
  People may associate relevant background knowledge according to current conversation, and then make their replies based on both context and knowledge.
  Recently, the tasks of knowledge-grounded response selection \cite{DBLP:conf/acl/KielaWZDUS18,DBLP:conf/emnlp/ZhouPB18} have been set up to simulate this scenario.
  In these tasks, agents should respond according to not only the given context but also the relevant knowledge, and the knowledge is usually represented as unstructured entries which are common in practice.
  An example is shown in Figure~\ref{fig1}.

  Some methods have been proposed for solving these tasks \cite{DBLP:conf/emnlp/MazareHRB18,DBLP:conf/ijcai/ZhaoTWX0Y19,DBLP:conf/emnlp/GuLZL19}.
  In these methods, the semantic representations of context, knowledge and responses candidates are usually derived by encoding models at first.
  Then, the matching degree between a response candidate and a \{context, knowledge\} pair is calculated by neural networks.
  Although these methods are capable of utilizing external knowledge when selecting responses, they still have several deficiencies.
  First, most of them encode context and knowledge separately, and neglect to ground the conversation on the knowledge and to comprehend the knowledge based on the conversation.
  \citet{DBLP:conf/ijcai/ZhaoTWX0Y19} proposed to alleviate this issue by fusing the \emph{local} matching information between each \{context utterance, knowledge entry\} pair into their representations.
  However, each utterance or entry plays different functions in conversations.
  As shown by the example in Figure~\ref{fig1}, some utterances are closely related with background knowledge while some others are irrelevant to knowledge but play the role of connection, such as the \emph{greetings}.
  Besides, some entries are redundant and are not mentioned in the conversation at all, such as \emph{Year}, \emph{Director} and \emph{Critical Response}.
  Such \emph{global} functions of utterances and entries were ignored in all existing methods.
  Second, the model structures used by previous methods to calculate the matching degree between a response candidate and a \{context, knowledge\} pair were usually \emph{shallow} ones, which constrained the model from learning \emph{deep} matching relationship between them.

  Therefore, this paper proposes a method named \textbf{F}iltering before \textbf{I}teratively \textbf{RE}ferring (\textbf{FIRE}) to address these issues.
  First, this method designs a context filter and a knowledge filter at the encoding stage.
  Different from \citet{DBLP:conf/ijcai/ZhaoTWX0Y19}, these filters collect the \emph{global} matching information between all context utterances and all knowledge entries bidirectionally.
  Specifically, the context filter makes the context refer to the knowledge and derives \emph{knowledge-aware context representations}.
  On the other hand, the knowledge filter derives \emph{context-aware knowledge representations} utilizing the same global attention mechanism.
  Considering that the knowledge entries are independent of each other and redundant entries may increase the difficulty of response matching, the knowledge filter discards irrelevant entries, which are determined by calculating the similarity between each entry and the whole context.

  Second, this method designs an iteratively referring network for calculating the matching degree between a response candidate and a \{context, knowledge\} pair.
  This network follows the dual matching framework \cite{DBLP:conf/emnlp/GuLZL19} in which the response refers to the context and the knowledge simultaneously.
  Motivated by previous studies on attention-over-attention (AoA) \cite{DBLP:conf/acl/CuiCWWLH17} and interaction-over-interaction (IoI) \cite{DBLP:conf/acl/TaoWXHZY19} models, this network performs the referring operation iteratively in order to derive \emph{deep} matching information.
  Specifically, the outputs of each iteration are utilized as the inputs of the next iteration.
  Then, the outputs of all iterations are aggregated into a set of matching feature vectors for scoring.

  We evaluate our proposed method on the PERSONA-CHAT  \cite{DBLP:conf/acl/KielaWZDUS18} and CMU\_DoG  \cite{DBLP:conf/emnlp/ZhouPB18} datasets.
  Experimental results show that FIRE outperforms previous methods by margins larger than 2.8\% and 4.1\% on the PERSONA-CHAT dataset with original and revised personas respectively, and margins larger than 3.1\% on the CMU\_DoG dataset in terms of top-1 accuracy, achieving a new state-of-the-art performance on both tasks.

  In summary, the contributions of this paper are three-fold.
  (1) A \textbf{F}iltering before \textbf{I}teratively \textbf{RE}ferring (\textbf{FIRE}) method is proposed, which employs two filtering structures based on global and cross attentions for representing contexts and knowledge, together with an iteratively referring network for scoring response candidates.
  (2) Experimental results on two datasets demonstrate that our proposed model outperforms state-of-the-art models on the accuracy of response selection.
  (3) Empirical analysis further verifies the effectiveness of our proposed method.

\section{Related Work}

  \subsection{Response Selection}
    Response selection is an important problem in building retrieval-based chatbots.
    Existing work on response selection can be categorized according to processing single-turn dialogues \cite{DBLP:conf/emnlp/WangLLC13} or multi-turn ones \cite{DBLP:conf/sigdial/LowePSP15,DBLP:conf/acl/WuWXZL17,DBLP:conf/coling/ZhangLZZL18,DBLP:conf/acl/WuLCZDYZL18,DBLP:conf/cikm/GuLL19,DBLP:conf/acl/TaoWXHZY19,DBLP:conf/cikm/GuLLLSWZ20,DBLP:journals/taslp/GuLL20}.
    Recent studies focused on multi-turn conversations, a more practical setup for real applications.
    \citet{DBLP:conf/acl/WuWXZL17} proposed the sequential matching network (SMN) which accumulated the utterance-response matching information by a recurrent neural network.
    \citet{DBLP:conf/acl/WuLCZDYZL18} proposed the deep attention matching network (DAM) to construct representations at different granularities with stacked self-attention.
    \citet{DBLP:conf/cikm/GuLL19} proposed the interactive matching network (IMN) to perform the bidirectional and global interactions between the context and the response.
    \citet{DBLP:conf/acl/TaoWXHZY19} proposed the interaction over interaction (IoI) model which performed matching by stacking multiple interaction blocks.
    \citet{DBLP:conf/cikm/GuLLLSWZ20} proposed the speaker-aware BERT (SA-BERT) to model the speaker change information in pre-trained language models.

  \subsection{Knowledge-Grounded Chatbots}

    Chit-chat models suffer from the lack of explicit long-term memory as they are typically trained to produce an utterance given only a very recent dialogue history.
    Recently, some studies show that chit-chat models can be more diverse and engaging by conditioning them on the background knowledge.
    \citet{DBLP:conf/acl/KielaWZDUS18} released the PERSONA-CHAT dataset which employs the speakers' profile information as the background knowledge. %, in order to maintain a consistent personality.
    \citet{DBLP:conf/emnlp/ZhouPB18} built the CMU\_DoG dataset which adopts the Wikipedia articles about popular movies as the background knowledge. %, in order to provide a relevant chat history and a source of information the models could use.
    \citet{DBLP:conf/emnlp/MazareHRB18} proposed to pre-train a model using a large-scale corpus based on Reddit.
    \citet{DBLP:conf/ijcai/ZhaoTWX0Y19} proposed the document-grounded matching network (DGMN) which fused each context utterance with each knowledge entry for representing them.
    \citet{DBLP:conf/emnlp/GuLZL19} proposed a dually interactive matching network (DIM) which performed the interactive matching between responses and contexts and between responses and knowledge respectively.

    The FIRE model proposed in this paper makes two major improvements to the state-of-the-art DIM model \cite{DBLP:conf/emnlp/GuLZL19}.
    First, a context filter and a knowledge filter are built to make the representations of context and knowledge aware of each other.
    Second, an iteratively referring network is designed to collect deep and comprehensive matching information for scoring responses.

\section{Task Definition}

  Given a dataset $\mathcal{D}$, an example is represented as $(c,k,r,y)$.
  Specifically, $c = \{u_1,u_2,...,u_{n_c}\}$ represents a context with $\{u_m\}_{m=1}^{n_c}$ as its utterances and $n_c$ as the utterance number.
  $k = \{e_1,e_2,...,e_{n_k}\}$ represents a knowledge description with $\{e_n\}_{n=1}^{n_k}$ as its entries and $n_k$ as the entry number.
  $r$ represents a response candidate.
  $y \in \{0,1\}$ denotes a label.
  $y=1$ indicates that $r$ is a proper response for $(c,k)$; otherwise, $y=0$.
  Our goal is to learn a matching model $g(c,k,r)$ from $\mathcal{D}$.
  For any context-knowledge-response triple $(c,k,r)$, $g(c,k,r)$ measures the matching degree between $(c,k)$ and $r$.

\section{FIRE Model}

  \begin{figure*}
    \centering
    \includegraphics[width=16cm]{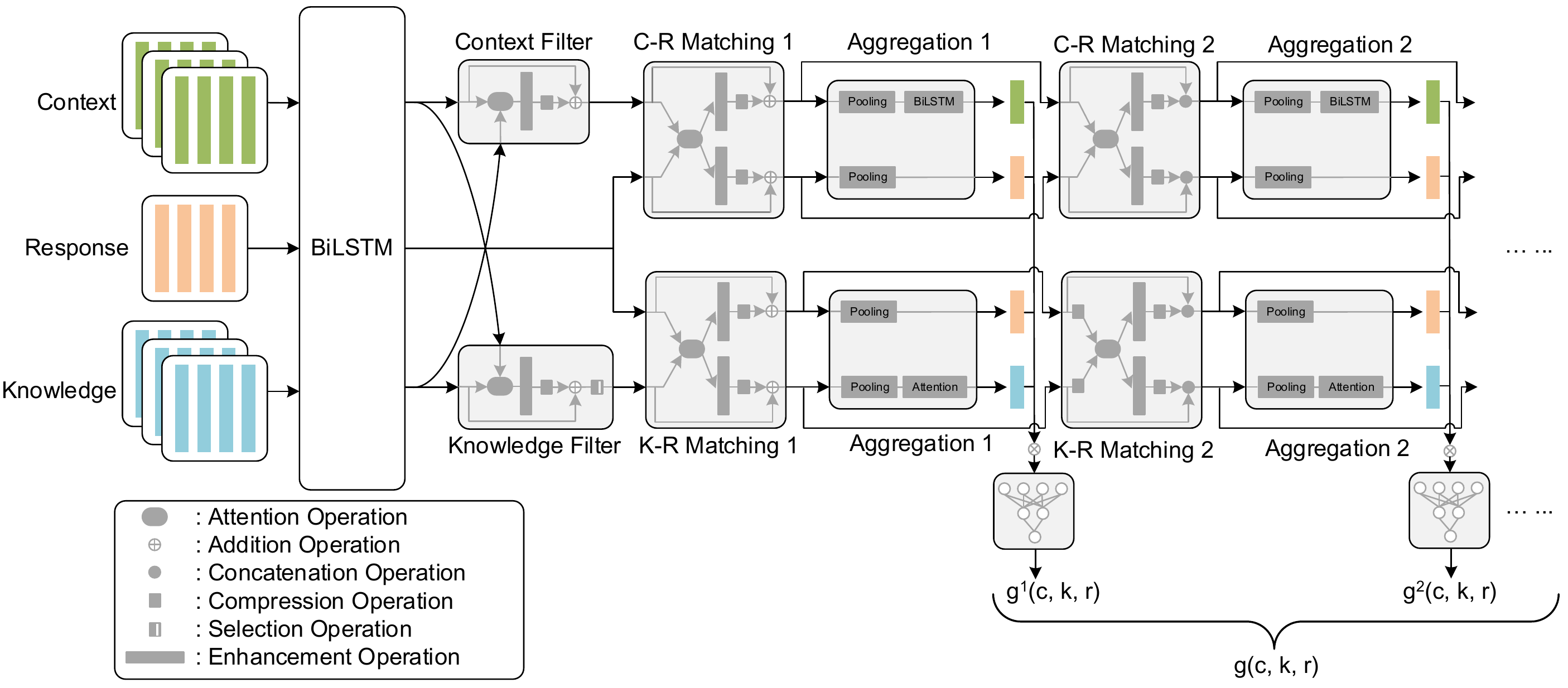}
    \caption{The overview architecture of our proposed FIRE model.}
    \label{fig2}
  \end{figure*}

    Figure~\ref{fig2} shows the overview architecture of our proposed model.
    The context utterances, knowledge entries and responses are first encoded by a sentence encoder.
    Then the context and the knowledge are co-filtered by referring to each other.
    Next, the response refers to the filtered context and knowledge representations iteratively.
    The outputs of each iteration are aggregated into a matching feature vector, and are utilized as the inputs of next iteration at the same time.
    Finally, the matching features of all iterations are accumulated for scoring response candidates.
    Details are provided in following subsections.

  \subsection{Word Representation}
    We follow the settings used in DIM \cite{DBLP:conf/emnlp/GuLZL19}, which constructs word representations by combining general pre-trained word embeddings,
    those estimated on the task-specific training set,
    as well as character-level embeddings, in order to deal with the out-of-vocabulary issue.

    Formally, embeddings of the \emph{m}-th utterance in a context, the \emph{n}-th entry in a knowledge description and a response candidate are denoted as
    $\textbf{U}_m = \{\textbf{u}_{m,i}\}_{i=1}^{l_{u_m}}$,
    $\textbf{E}_n = \{\textbf{e}_{n,j}\}_{j=1}^{l_{e_n}}$ and
    $\textbf{R} = \{\textbf{r}_k\}_{k=1}^{l_r}$ respectively,
    where $l_{u_m}$, $l_{e_n}$ and $l_r$ are the numbers of words in $\textbf{U}_m$, $\textbf{E}_n$ and $\textbf{R}$ respectively.
    Each $\textbf{u}_{m,i}$, $\textbf{e}_{n,j}$ or $ \textbf{r}_k$ is an embedding vector.

  \subsection{Sentence Encoder} \label{sec6}
    Note that the encoder can be any existing encoding model.
    In this paper, the context utterances, knowledge entries and response candidate are encoded by bidirectional long short-term memories (BiLSTMs) \cite{DBLP:journals/neco/HochreiterS97}.
    Detailed calculations are omitted due to limited space.
    After that, we can obtain the encoded representations for utterances, entries and response, denoted as
    $\bar{\textbf{U}}_m = \{\bar{\textbf{u}}_{m,i}\}_{i=1}^{l_{u_m}}$,
    $\bar{\textbf{E}}_n = \{\bar{\textbf{e}}_{n,j}\}_{j=1}^{l_{e_n}}$ and
    $\bar{\textbf{R}} = \{\bar{\textbf{r}}_k\}_{j=1}^{l_r}$ respectively.
    Each $\bar{\textbf{u}}_{m,i}$, $\bar{\textbf{e}}_{n,j}$ or $\bar{\textbf{r}}_k$ is an embedding vector of \emph{d}-dimensions.
    The parameters of these three BiLSTMs are shared in our implementation.

  \subsection{Context and Knowledge Filters}
    As illustrated in Figure~\ref{fig1}, not every context utterance refers to the knowledge, and not every knowledge entry is mentioned in the conversation.
    In order to ground the conversation on the knowledge and to comprehend the knowledge based on the conversation, we build a context filter and a knowledge filter in the FIRE model.
    These two filters obtain \emph{knowledge-aware context representation} $\bar{\textbf{C}}^0$ and \emph{context-aware knowledge representation} $\bar{\textbf{K}}^0$, which are further utilized to match with the response.

    \paragraph{Context Filter}
    This filter first determines the knowledge that each context token refers to by a \emph{global} attention between the whole context and all knowledge entries.
    Then, it enhances the representation of each context token with the representations of its relevant knowledge.

    Given the set of utterance representations $\{\bar{\textbf{U}}_m\}_{m=1}^{n_c}$ encoded by the sentence encoder, we concatenate them to form the context representation $\bar{\textbf{C}} = \{\bar{\textbf{c}}_i\}_{i=1}^{l_c} $ with $l_c = \sum_{m=1}^{n_c} l_{u_m}$.
    Also, the knowledge representation $\bar{\textbf{K}} = \{\bar{\textbf{k}}_j\}_{j=1}^{l_k} $ with $l_k = \sum_{n=1}^{n_k} l_{e_n}$ is formed similarly by concatenating $\{\bar{\textbf{E}}_n\}_{n=1}^{n_k}$.
    Then, a soft alignment is performed by computing the attention weight between each tuple \{$\bar{\textbf{c}}_i, \bar{\textbf{k}}_j$\} as
    \begin{align}
      e_{ij} = \bar{\textbf{c}}_i^\top \cdot \bar{\textbf{k}}_j.
    \end{align}
    After that, the global relevance between the context and the knowledge can be obtained using these attention weights.
    For a word in the context, its relevant representation carried by the knowledge is identified and composed using $e_{ij}$ as
    \begin{align}
      \tilde{\textbf{c}}_i = \sum_{j=1}^{l_k} \frac{exp(e_{ij})} {\sum_{z=1}^{l_k} exp(e_{iz})} \bar{\textbf{k}}_j, i \in \{1, ..., l_c\},
      \label{equ1}
    \end{align}
    where the contents in $\{\bar{\textbf{k}}_j\}_{j=1}^{l_k}$ that are relevant to $\bar{\textbf{c}}_i$ are selected to form $\tilde{\textbf{c}}_i$, and we define $\tilde{\textbf{C}} = \{\tilde{\textbf{c}}_i\}_{i=1}^{l_c}$.

    To enhance the context representation $\bar{\textbf{C}}$ with the relevance representation $\tilde{\textbf{C}}$, the element-wise difference and multiplication between \{$\bar{\textbf{C}}, \tilde{\textbf{C}}$\} are computed, and are then concatenated with the original vectors.
    This enhancement operation can be written as
    \begin{align}
      \widehat{\textbf{C}} = [\bar{\textbf{C}}; \tilde{\textbf{C}}; \bar{\textbf{C}} - \tilde{\textbf{C}} ;\bar{\textbf{C}} \odot \tilde{\textbf{C}}],
    \end{align}
    where $\widehat{\textbf{C}} = \{\hat{\textbf{c}}_i\}_{i=1}^{l_c}$ and $\hat{\textbf{c}}_i \in \mathbb{R}^{4d}$.
    Finally, we compress $\widehat{\textbf{C}}$ and obtain the {knowledge-aware context} representation $\bar{\textbf{C}}^0$ as
    \begin{align}
      \bar{\textbf{c}}_i^0 &= \textbf{ReLU}( \hat{\textbf{c}}_i \cdot \textbf{W}_c + \textbf{b}_c ) + \bar{\textbf{c}}_i,
    \end{align}
    where $\bar{\textbf{C}}^0 = \{\bar{\textbf{c}}_i^0\}_{i=1}^{l_c}$, $ \textbf{W}_c \in \mathbb{R}^{4d \times d} $ and $ \textbf{b}_c \in \mathbb{R}^{d} $ are parameters updated during training.

    Here, we define a referring function to summarize above operations in the context filter as
    \begin{align}
      \bar{\textbf{C}}^0 = \textbf{REFER} (\bar{\textbf{C}}, \bar{\textbf{K}}),
    \end{align}
    where $\bar{\textbf{C}}$ acts as the \emph{query}, and $\bar{\textbf{K}}$ acts as the \emph{key} and \emph{value} of the referring function respectively.

    \paragraph{Knowledge Filter}
    Similarly, this filter enhances the representation of each knowledge token with the representations of its relevant context.
    Different from the context filter, an additional selection operation is conducted to directly filter out the knowledge entries with low relevance with the context since the entries are independent of each other.

    First, the referring function introduced above is also performed as follows,
    \begin{align}
      \bar{\textbf{K}}^{0'} = \textbf{REFER} (\bar{\textbf{K}}, \bar{\textbf{C}}).
    \end{align}
    where $\bar{\textbf{K}}^{0'}$ is the {context-aware knowledge} representation and $\bar{\textbf{K}}^{0'} = \{\bar{\textbf{E}}_n^{0'}\}_{n=1}^{n_k}$.

    Furthermore, the relevance between each entry and the whole conversation is computed in order to determine whether to filter out this entry.
    We first perform the last-hidden-state pooling over the representations of utterances and entries given by the sentence encoder in Section~\ref{sec6}.
    Then, the utterance embedding $\{\bar{\textbf{u}}_m\}_{m=1}^{n_c}$  and the entry embedding $\{\bar{\textbf{e}}_n\}_{n=1}^{n_k}$ are obtained.
    Next, we compute the relevance score for each utterance-entry pair as follows,
    \begin{align}
      s_{mn} =  \bar{\textbf{u}}_m^\top \cdot \textbf{M} \cdot \bar{\textbf{e}}_n,
      \label{equ2}
    \end{align}
    where $\textbf{M} \in \mathbb{R}^{d \times d}$ is a matrix that needs to be estimated.

    In order to obtain the overall relevance score between each entry and the whole conversation, an aggregation operation is required.
    Here, we make an assumption that one entry is mentioned only once in the conversation.
    Thus, for a given entry, its relevance score with the conversation is defined as the maximum relevance score between it and all utterances.
    Mathematically, we have
    \begin{align}
      s_n    &= \max \limits_{m} \ s_{mn}.
      \label{equ3}
    \end{align}
    Those entries whose scores are below a threshold $\gamma$ are considered as uninformative ones for the conversation and are directly filtered out before matching with responses.
    Mathematically, we have
    \begin{align}
      \bar{\textbf{E}}_n^0 &= \max(0,sgn(\sigma(s_n)-\gamma)) \cdot \bar{\textbf{E}}_n^{0'}, % n \in \{1, ..., n_k\},
    \end{align}
    where $\sigma$ is the sigmoid function and $sgn$ is the sign function.
    The final filtered knowledge representation is defined as $\bar{\textbf{K}}^0 = \{ \bar{\textbf{E}}_n^0\}_{n=1}^{n_k}$.

  \subsection{Iteratively Referring}

    \citet{DBLP:conf/ijcai/ZhaoTWX0Y19} and \citet{DBLP:conf/emnlp/GuLZL19} showed that the referring operation between contexts and responses and that between knowledge and responses can both provide useful matching information for response selection.
    However, the matching information collected by these methods were very \textit{shallow} and \textit{limited}, as each response candidate referred to the context or the knowledge only once in their models.
    In this paper, we design an iteratively referring network which makes the response refer to the filtered context and knowledge iteratively.
    Each iteration is capable of capturing additional matching information based on previous ones.
    Accumulating these iterations can help to derive the \textit{deep} and \textit{comprehensive} matching features for response selection.

    Take the context-response matching as an example.
    The matching strategy adopted here considers the global and bidirectional matching between two sequences.
    Let $\bar{\textbf{C}}^l = \{\bar{\textbf{c}}_i^l\}_{i=1}^{l_c}$ and $\bar{\textbf{R}}^l = \{\bar{\textbf{r}}_k^l\}_{k=1}^{l_r}$ be the outputs of the \textit{l}-th iteration, i.e., the inputs of the (\textit{l+1})-th iteration,
    where $l \in \{0, 1, ..., L-1\}$ and $L$ is the number of iterations.
    For response representations, we have $\bar{\textbf{R}}^0 = \bar{\textbf{R}}$.

    First, the context refers to the response by performing the referring function and the \emph{response-aware context representation}  $\bar{\textbf{C}}^{l+1}$ is obtained as
    \begin{align}
      \bar{\textbf{C}}^{l+1} = \textbf{REFER} (\bar{\textbf{C}}^l, \bar{\textbf{R}}^l).
    \end{align}
    Bidirectionally, the response refers to the context and the \emph{context-aware response representation} $\bar{\textbf{R}}^{l+1}$ is obtained as
    \begin{align}
      \bar{\textbf{R}}^{l+1} = \textbf{REFER} (\bar{\textbf{R}}^l, \bar{\textbf{C}}^l).
    \end{align}
    $\bar{\textbf{C}}^{l+1}$ and $\bar{\textbf{R}}^{l+1}$ are utilized as the input of next iteration.
    Finally, $\{\bar{\textbf{C}}^l\}_{l=1}^L$ and $\{\bar{\textbf{R}}^l\}_{l=1}^L$ are obtained after \textit{L} iterations.

    On the other hand, the knowledge-response matching is conducted identically to the context-response matching process introduced above.
    The \emph{response-aware knowledge representation} $\bar{\textbf{K}}^l$ and \emph{knowledge-aware response representation} $\bar{\textbf{R}}^{l*}$ are iteratively updated as
    \begin{align}
      \bar{\textbf{K}}^{l+1} = \textbf{REFER} (\bar{\textbf{K}}^l, \bar{\textbf{R}}^{l*}), \\
      \bar{\textbf{R}}^{l+1*} = \textbf{REFER} (\bar{\textbf{R}}^{l*}, \bar{\textbf{K}}^l),
    \end{align}
    where $\bar{\textbf{R}}^{0*} = \bar{\textbf{R}}$.
    Similarly, we  obtain
    $\{\bar{\textbf{K}}^l\}_{l=1}^L$ and $\{\bar{\textbf{R}}^{l*}\}_{l=1}^L$ after \textit{L} iterations.

  \subsection{Aggregation}
    These sets of matching matrices
    $\{\bar{\textbf{C}}^l\}_{l=1}^L$, $\{\bar{\textbf{R}}^l\}_{l=1}^L$,
    $\{\bar{\textbf{K}}^l\}_{l=1}^L$, and $\{\bar{\textbf{R}}^{l*}\}_{l=1}^L$
    are aggregated into a set of matching feature vectors finally.
    As shown in Figure~\ref{fig1}, we perform the same aggregation operation after each referring iteration.
    The aggregation strategy in DIM \cite{DBLP:conf/emnlp/GuLZL19} is adopted here.

    Let us take the $l$-th aggregation as an example.
    First, $\bar{\textbf{C}}^l$ and $\bar{\textbf{K}}^l$ are converted back to the matching matrices  $\{\bar{\textbf{U}}_m^l\}_{m=1}^{n_c}$ and $\{\bar{\textbf{E}}_n^l\}_{n=1}^{n_k}$ for separate utterances and entries.
    Then, each matching matrix $\bar{\textbf{U}}_m^l, \bar{\textbf{R}}^l, \bar{\textbf{E}}_n^l$, and $\bar{\textbf{R}}^{l*}$ are aggregated by max pooling and mean pooling operations to derive their embedding vectors $\bar{\textbf{u}}_m^l$, $\bar{\textbf{r}}^l$, $\bar{\textbf{e}}_n^l$ and $\bar{\textbf{r}}^{l*}$ respectively.
    Next, the sequences of $\{\bar{\textbf{u}}_m^l\}_{m=1}^{n_c}$ and $\{\bar{\textbf{e}}_n^l\}_{n=1}^{n_k}$ are further aggregated to get the embedding vectors for the context and the knowledge respectively.

    As the utterances in a context are chronologically ordered, the utterance embeddings $\{\bar{\textbf{u}}_m^l\}_{m=1}^{n_c}$ are sent into another BiLSTM following the chronological order of utterances in the context.
    Combined max pooling and last-hidden-state pooling operations are then performed to derive the context embeddings $\bar{\textbf{c}}^l$.
    On the other hand, as the knowledge entries are independent of each other, an attention-based aggregation is designed to derive the knowledge embeddings $\bar{\textbf{k}}^l$. Readers can refer to \citet{DBLP:conf/emnlp/GuLZL19} for more details.

    The matching feature vector of the $l$-th iteration is the concatenation of context, knowledge and response embeddings as
    \begin{align}
      \textbf{m}^l = [\bar{\textbf{c}}^l;\bar{\textbf{r}}^l;\bar{\textbf{k}}^l;\bar{\textbf{r}}^{l*}],
    \end{align}
    which combines the outputs of both context-response matching and knowledge-response matching.

    Last, we obtain a set of matching feature vectors $\{\textbf{m}^l\}_{l=1}^L$ for all iterations.

  \subsection{Prediction}
    Each matching feature vector $\textbf{m}^l$ is sent into a multi-layer perceptron (MLP) classifier.
    Here, the MLP is designed to predict the matching degree $g^l(c,k,r)$ between $r$ and $(c,k)$ at $l$-th iteration.
    A softmax output layer is adopted in the MLP to return a probability distribution over all response candidates.
    The probability distributions calculated from all $L$ matching feature vectors are averaged to derive the final distribution for ranking.

  \subsection{Model Learning}
    Inspired by \citet{DBLP:conf/acl/TaoWXHZY19}, the model parameters of FIRE are learnt by minimizing the summation of cross-entropy losses of MLP at all iterations.
    By this means, each matching feature vector can be directly supervised by labels in the training set.
    Furthermore, inspired by \citet{DBLP:conf/cvpr/SzegedyVISW16}, we employ the strategy of label smoothing by assigning a small additional confidence $\epsilon$ to all candidates, in order to prevent the model from being over-confident.
    Let $\Theta$ denote the parameters of FIRE.
    The learning objective $\mathcal{L}(\mathcal{D}, \Theta)$ is formulated as
    \begin{equation}
      %\small
      \mathcal{L}(\mathcal{D}, \Theta) = - \sum_{l=1}^L \sum_{(c,k,r,y)\in \mathcal{D}} (y+\epsilon) log(g^l(c,k,r)) .
    \end{equation}

\section{Experiments}

    \begin{table*}[!hbt]
    %   \small
      \centering
      \setlength{\tabcolsep}{1.3mm}{
      \begin{tabular}{l|c|c|c|c|c|c|c|c|c}
      \toprule
      \multirow{3}*{Model}                                &    \multicolumn{6}{c|}{PERSONA-CHAT}    & \multicolumn{3}{c}{\multirow{2}*{CMU\_DoG}}  \\
      \cline{2-7}
                                                          &  \multicolumn{3}{c|}{Original} & \multicolumn{3}{c|}{Revised}  \\
      \cline{2-10}
                                                          & $\textbf{R}@1$ & $\textbf{R}@2$ & $\textbf{R}@5$ & $\textbf{R}@1$ & $\textbf{R}@2$ & $\textbf{R}@5$ & $\textbf{R}@1$ & $\textbf{R}@2$ & $\textbf{R}@5$ \\
      \hline
      Starspace \cite{DBLP:conf/aaai/WuFCABW18}           & 49.1 & 60.2 & 76.5 & 32.2 & 48.3 & 66.7 & 50.7 & 64.5 & 80.3 \\
      Profile Memory \cite{DBLP:conf/acl/KielaWZDUS18}    & 50.9 & 60.7 & 75.7 & 35.4 & 48.3 & 67.5 & 51.6 & 65.8 & 81.4 \\
      KV Profile Memory \cite{DBLP:conf/acl/KielaWZDUS18} & 51.1 & 61.8 & 77.4 & 35.1 & 45.7 & 66.3 & 56.1 & 69.9 & 82.4 \\
      Transformer \cite{DBLP:conf/emnlp/MazareHRB18}      & 54.2 & 68.3 & 83.8 & 42.1 & 56.5 & 75.0 & 60.3 & 74.4 & 87.4 \\
      DGMN \cite{DBLP:conf/ijcai/ZhaoTWX0Y19}             & 67.6 & 80.2 & 92.9 & 58.8 & 62.5 & 87.7 & 65.6 & 78.3 & 91.2 \\
      DIM \cite{DBLP:conf/emnlp/GuLZL19}                  & 78.8 & 89.5 & 97.0 & 70.7 & 84.2 & 95.0 & 78.7 & 89.0 & 97.1 \\
      \hline
      FIRE (Ours)                                         & \textbf{81.6} & \textbf{91.2} & \textbf{97.8} & \textbf{74.8} & \textbf{86.9} & \textbf{95.9} & \textbf{81.8} & \textbf{90.8} &	\textbf{97.4} \\
      \bottomrule
      \end{tabular}}
      \caption{Performance of FIRE and previous methods on the test sets of PERSONA-CHAT and CMU\_DoG datasets.
               The meanings of ``Original", and ``Revised" can be found in Section~\ref{sec1}.}
      \label{tab1}
    \end{table*}

  \subsection{Datasets} \label{sec1}
    We tested our proposed method on the PERSONA-CHAT \cite{DBLP:conf/acl/KielaWZDUS18} and CMU\_DoG \cite{DBLP:conf/emnlp/ZhouPB18} datasets which both contain dialogues grounded on background knowledge.

    The PERSONA-CHAT dataset consists of 8939 complete dialogues for training, 1000 for validation, and 968 for testing.
    Response selection is performed at every turn of a complete dialogue, which results in 65719 dialogues for training, 7801 for validation, and 7512 for testing in total.
    Positive responses are true responses from humans and negative ones are randomly sampled by the dataset publishers.
    The ratio between positive and negative responses is 1:19 in the training, validation, and testing sets.
    There are 955 personas for training, 100 for validation, and 100 for testing, each consisting of 3 to 5 profile sentences.
    To make this task more challenging, a version of revised persona descriptions are provided by rephrasing, generalizing, or specializing the original ones.

    The CMU\_DoG dataset consists of 2881 complete dialogues for training, 196 for validation, and 537 for testing.
    Response selection is also performed at every turn of a complete dialogue, which results in 36159 dialogues for training, 2425 for validation, and 6637 for testing in total.
%    This dataset was built in two scenarios.
%    In the first scenario, only one worker had access to the provided knowledge, and he/she was responsible for introducing the movie to the other worker.
%    While in the second scenario, both workers knew the knowledge and they were asked to discuss the content.
%    Since the data size for each scenario is small, we followed the setting used in \citet{DBLP:conf/ijcai/ZhaoTWX0Y19} which merged the data of the two scenarios for experiments and filtered out the conversations with less than 4 turns to avoid noise.
    Since this dataset did not contain negative examples, we adopted the version shared by \citet{DBLP:conf/ijcai/ZhaoTWX0Y19}, in which 19 negative candidates were randomly sampled for each utterance from the same set.

  \subsection{Evaluation Metrics}
    We used the same evaluation metrics as the ones in  previous work \cite{DBLP:conf/acl/KielaWZDUS18,DBLP:conf/ijcai/ZhaoTWX0Y19}.
    Each model aimed to select $k$ best-matched response from available candidates for the given context and knowledge.
    Then, the recall of true positive replies, denoted as $\textbf{R}@k$, are calculated as the measurement.

  \subsection{Training Details}

    For training FIRE on both PERSONA-CHAT and CMU\_DoG datasets, some common configurations were set as follows.
    The Adam method \cite{DBLP:journals/corr/KingmaB14} was employed for optimization.
    The  learning rate was initialized as 0.00025 and was exponentially decayed by 0.96 every 5000 steps.
    Dropout \cite{DBLP:journals/jmlr/SrivastavaHKSS14} with a rate of 0.2 was applied to the word embeddings and all hidden layers.
    The word representation was the concatenation of a 300-dimensional GloVe embedding \cite{DBLP:conf/emnlp/PenningtonSM14}, a 100-dimensional embedding estimated on the training set using the Word2Vec algorithm \cite{DBLP:conf/nips/MikolovSCCD13}, and a 150-dimensional character-level embedding estimated by a CNN network that consists of 50 filters and window sizes were set to \{3, 4, 5\} respectively.
    The word embeddings were not updated during training.
    All hidden states of LSTMs had 200 dimensions.
    The MLP at the prediction layer had 256 hidden units with ReLU \cite{DBLP:conf/icml/NairH10} activation.
    $\epsilon$ used in label smoothing was set to 0.05.
    The validation set was used to select the best model for testing.

    Some configurations were different according to the characteristics of these two datasets.
    For the PERSONA-CHAT dataset, the maximum number of characters in a word, that of words in a context utterance, of utterances in a context, of words in a response, of words in a knowledge entry, and of entries in a knowledge description were set as 18, 20, 15, 20, 15, and 5 respectively.
    For the CMU\_DoG dataset, these parameters were set as 18, 40, 15, 40, 40 and 20 respectively.
    Zero-padding was adopted if the number of utterances in a context and the number of knowledge entries in a knowledge description were less than the maximum.
    Otherwise, we kept the last context utterances or the last knowledge entries.
    Batch size was set to 16 for PERSONA-CHAT and 4 for CMU\_DoG.
    The hyper-parameter $\gamma$ was set to 0.3 for original personas and 0.2 for revised personas on the PERSONA-CHAT dataset, as well as 0.2 on the CMU\_DoG dataset, which were tuned on the validation sets as shown in Figure~\ref{fig4}.
    The number of iterations $L$ was set to 3 for original and revised personas on the PERSONA-CHAT dataset, as well as 3 on the CMU\_DoG dataset, which were tuned on the validation sets as shown in Figure~\ref{fig5}.

    All code was implemented in the TensorFlow framework \cite{DBLP:conf/osdi/AbadiBCCDDDGIIK16} and is published to help replicate our results.\footnote{https://github.com/JasonForJoy/FIRE}

  \subsection{Experimental Results}

    Table~\ref{tab1} presents the evaluation results of FIRE and previous methods on the PERSONA-CHAT using original or revised personas and on the CMU\_DoG dataset.
    Because the paper proposing DIM \cite{DBLP:conf/emnlp/GuLZL19} only studied the PERSONA-CHAT dataset, we ran its released code to get the performance of DIM on the CMU\_DoG dataset.

    From Table~\ref{tab1}, we can see that FIRE achieved higher top-1 accuracy $\textbf{R}@1$ than all previous methods on both datasets, achieving a new state-of-the-art performance.
    On the PERSONA-CHAT dataset, the margins were larger than 2.8\% and 4.1\% when original and revised personas were used respectively.
    On the CMU\_DoG dataset, the margin was larger than 3.1\%. % on the CMU\_DoG dataset in terms of top-1 accuracy $\textbf{R}@1$, achieving a new state-of-the-art performance for knowledge-grounded response selection in retrieval-based chatbots.

  \subsection{Analysis}

    \paragraph{Ablation tests}

      \begin{table}[t]
      %   \small
        \centering
        \setlength{\tabcolsep}{0.9mm}{
        \begin{tabular}{l|c|c|c}
        \toprule
        \multirow{3}*{Model}       &    \multicolumn{2}{c|}{PERSONA-CHAT}    & \multicolumn{1}{c}{\multirow{2}*{CMU\_DoG}}  \\
        \cline{2-3}
                                   & Original & Revised  \\
        \cline{2-4}
                                   & $\textbf{R}@1$ & $\textbf{R}@1$ & $\textbf{R}@1$ \\
        \hline
        FIRE                       & 82.3 & 75.2 & 83.4 \\
        \ \ - Ite. Ref.            & 81.3 & 73.8 & 81.6 \\
        \ \ \ \ - Filters          & 78.9 & 71.1 & 78.8 \\
        \hline
        C-R                        & 65.6 & 66.2 & 79.7 \\
        C-R $\rightarrow$  Fusion  & 67.0 & 66.4 & 80.9 \\
        Filter $\rightarrow$ C-R   & 78.8 & 70.2 & 81.4 \\
        \hline
        K-R                        & 51.6 & 34.3 & 57.8 \\
        K-R $\rightarrow$  Fusion  & 54.2 & 39.4 & 63.1 \\
        Filter $\rightarrow$ K-R   & 63.6 & 51.0 & 73.5 \\
        \bottomrule
        \end{tabular}}
        \caption{The results of ablation tests on the validation sets. Here,  C-R denotes context-response matching and K-R denotes knowledge-response matching. The symbol $\rightarrow$ indicates the order of operations.}
        \label{tab2}
      \end{table}

      We conducted ablation tests as follows.
      First, we removed iteratively referring by setting the number of iterations $L$ to one.
      Then, we removed the two filters.
      The results on the validation sets are shown in Table~\ref{tab2}.
      We can see  the drop of $\textbf{R}@1$ after each step, which demonstrated the effectiveness of both components in FIRE.

      To further verify the effectiveness of the context filter, we built three models as follows:
      (1) a model that only performed the context-response matching without using any knowledge, i.e., the IMN model in \citet{DBLP:conf/emnlp/GuLZL19} where readers can refer to for more details;
      (2) a model that performed the context-response matching first and then fuse the knowledge, i.e., the IMN$_{utr}$ model in \citet{DBLP:conf/emnlp/GuLZL19}; and
      (3) a model that filtered the context first and then performed the context-response matching, i.e., our FIRE model with only the upper branch in Figure~\ref{fig2}.
      The evaluation results of these three models on the validation set are shown in Table~\ref{tab2}.
      Since these three models adopted similar context-response matching strategy, we can see that fusion after matching and filtering before matching can both improve the performance of response selection after introducing knowledge.
      Furthermore, filtering before matching outperformed fusion after matching by a large margin, which demonstrated the effectiveness of the context filter.
      On the other hand, we also built similar models to further verify the effectiveness of the knowledge filter.
      The same comparison results were observed from the last three rows of Table~\ref{tab2}, which demonstrated its effectiveness.

    \paragraph{Case Study}

      \begin{table}[t]
        %\small
        \centering
        \setlength{\tabcolsep}{1.2mm}{
        \begin{tabular}{l|l}
        \toprule
        \multicolumn{2}{l}{\textbf{Context Utterances}} \\
        \midrule
        \textbf{U1}  & hey , are you a student , i traveled a lot ,    \\
                     & i even studied abroad . \\
        \textbf{U2}  & no , i work full time at a nursing home .  \\
                     & i am a nurses aide . \\
        \textbf{U3}  & nice , i just got a advertising job myself .  \\
                     & do you like your job ? \\
        \textbf{U4}  & nice . yes i do . caring for people is the joy    \\
                     & of my life . \\
        \textbf{U5}  & nice my best friend is a nurse , i knew him  \\
                     & since kindergarten . \\
        \midrule
        \multicolumn{2}{l}{\textbf{Knowledge Entries}} \\
        \midrule
        \textbf{E1}& i have two dogs and one cat .  \\
        \textbf{E2}& i work as a nurses aide in a nursing home .    \\
        \textbf{E3}& i love to ride my bike .  \\
        \textbf{E4}& i love caring for people .    \\
        \bottomrule
        \end{tabular}}
        \caption{Context utterances and knowledge entries of a sample in the test set of the PERSONA-CHAT dataset.}
        \label{tab5}
      \end{table}

      \begin{figure}[t]
        \centering
        \subfigure[$s_{mn}$ in Eq.~(\ref{equ2})]{
        \includegraphics[width=4.5cm]{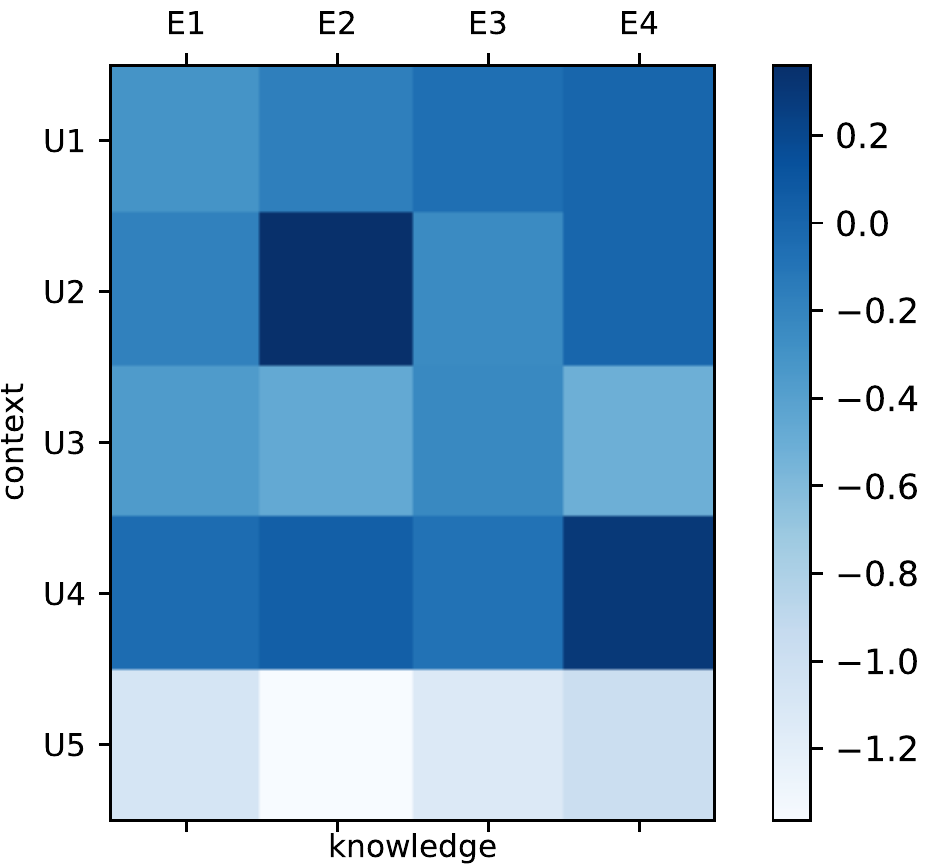}}
        \subfigure[$s_{n}$ in Eq.~(\ref{equ3})]{
        \includegraphics[width=4.5cm]{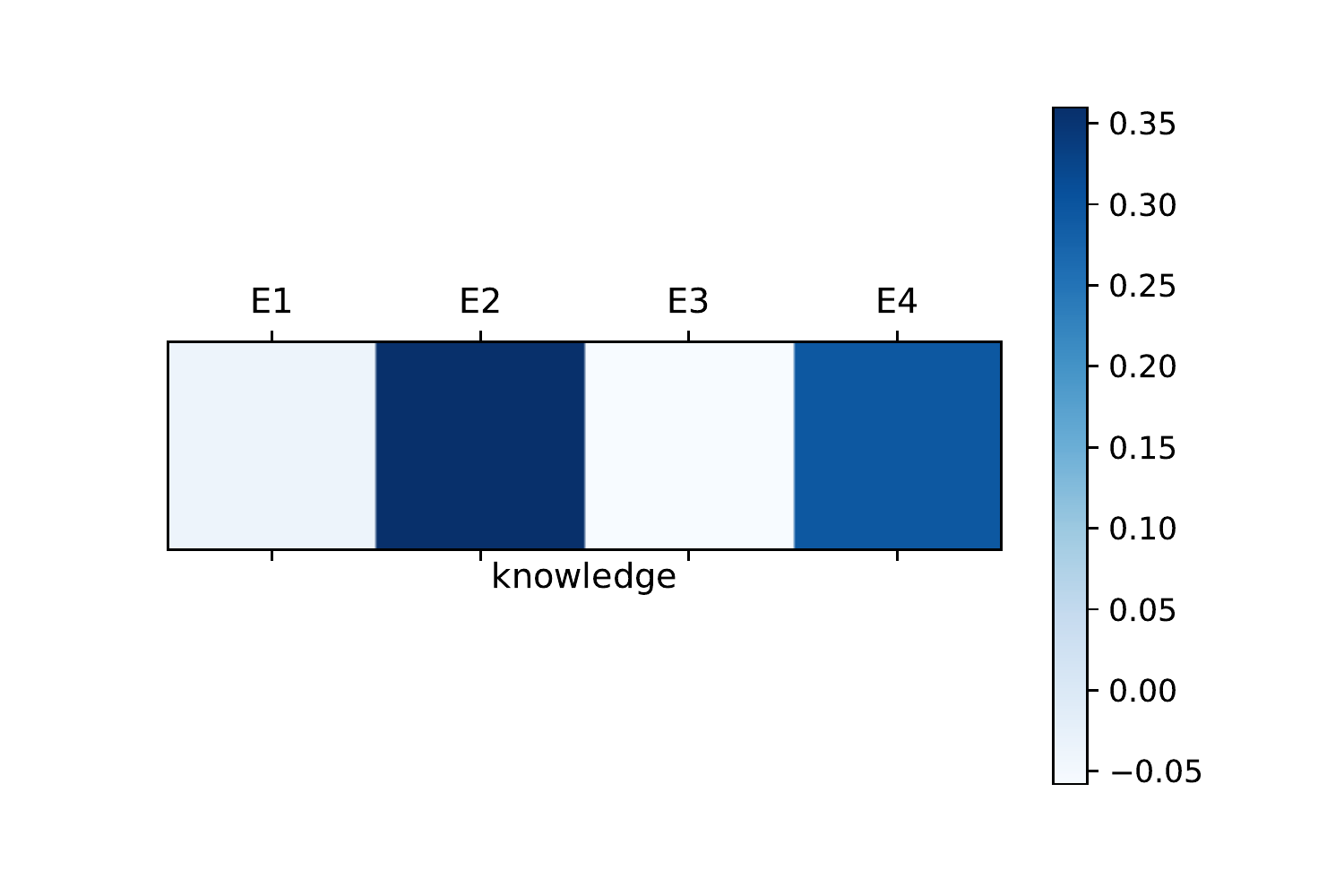}}
        \caption{Visualizations of  (a) $s_{mn}$ in Eq.~(\ref{equ2}) and (b) $s_{n}$ in Eq.~(\ref{equ3}) for a test sample of PERSONA-CHAT. The darker units correspond to larger values.}
        \label{fig6}
      \end{figure}

      A case study was conducted to visualize the attention weights in both context and knowledge filters of FIRE model.
      A sample was used as shown in Table~\ref{tab5}.
      The similarity scores $s_{mn}$ in Eq.~(\ref{equ2}) for each utterance-entry pair are visualized in Figure~\ref{fig6} (a).
      The final scores $s_{n}$ in Eq.~(\ref{equ3}) for each entry are visualized in Figure~\ref{fig6} (b).

      We can see that \textbf{U2} and \textbf{U4} obtained large attention weights with \textbf{E2} and \textbf{E4} respectively.
      Meanwhile, some irrelevant entries \textbf{E1} and \textbf{E3} obtained small similarity scores with the conversation, which can be filtered out with appropriate threshold.
      These experimental results verified the effectiveness of the filtering process and the interpretability of the knowledge grounding process.

    \paragraph{Knowledge Selection}

      \begin{figure}[t]
        \centering
        \includegraphics[width=7cm]{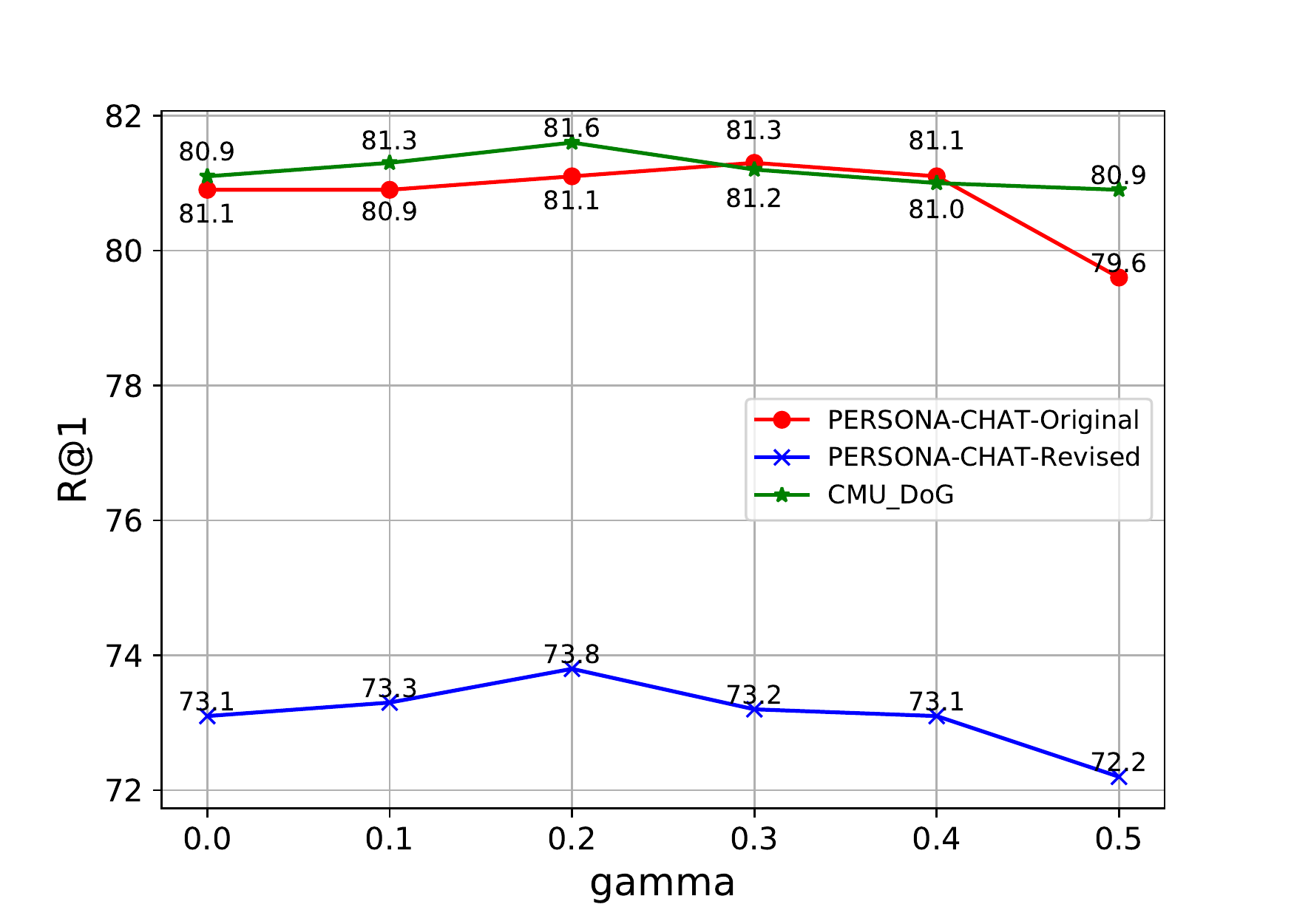}
        \caption{Validation set performance of FIRE with different threshold $\gamma$ in the knowledge filter.}
        \label{fig4}
      \end{figure}

      Figure~\ref{fig4} illustrates the validation set performance of FIRE with different threshold $\gamma$ in the knowledge filter.
      Here, the number of iterations $L$ was set to 1 for saving computation.
      When $\gamma=0$, no knowledge entries were filtered out.
      From this figure, we can observe a consistent trend that the model performance was improved when increasing $\gamma$ at the beginning, which indicates that filtering out irrelevant entries indeed helped response selection.
      Then, the performance started to drop when $\gamma$ was too large since some indeed relevant entries may be filtered out by mistake.

    \paragraph{Iteratively Referring}

      \begin{figure}[t]
        \centering
        \includegraphics[width=7cm]{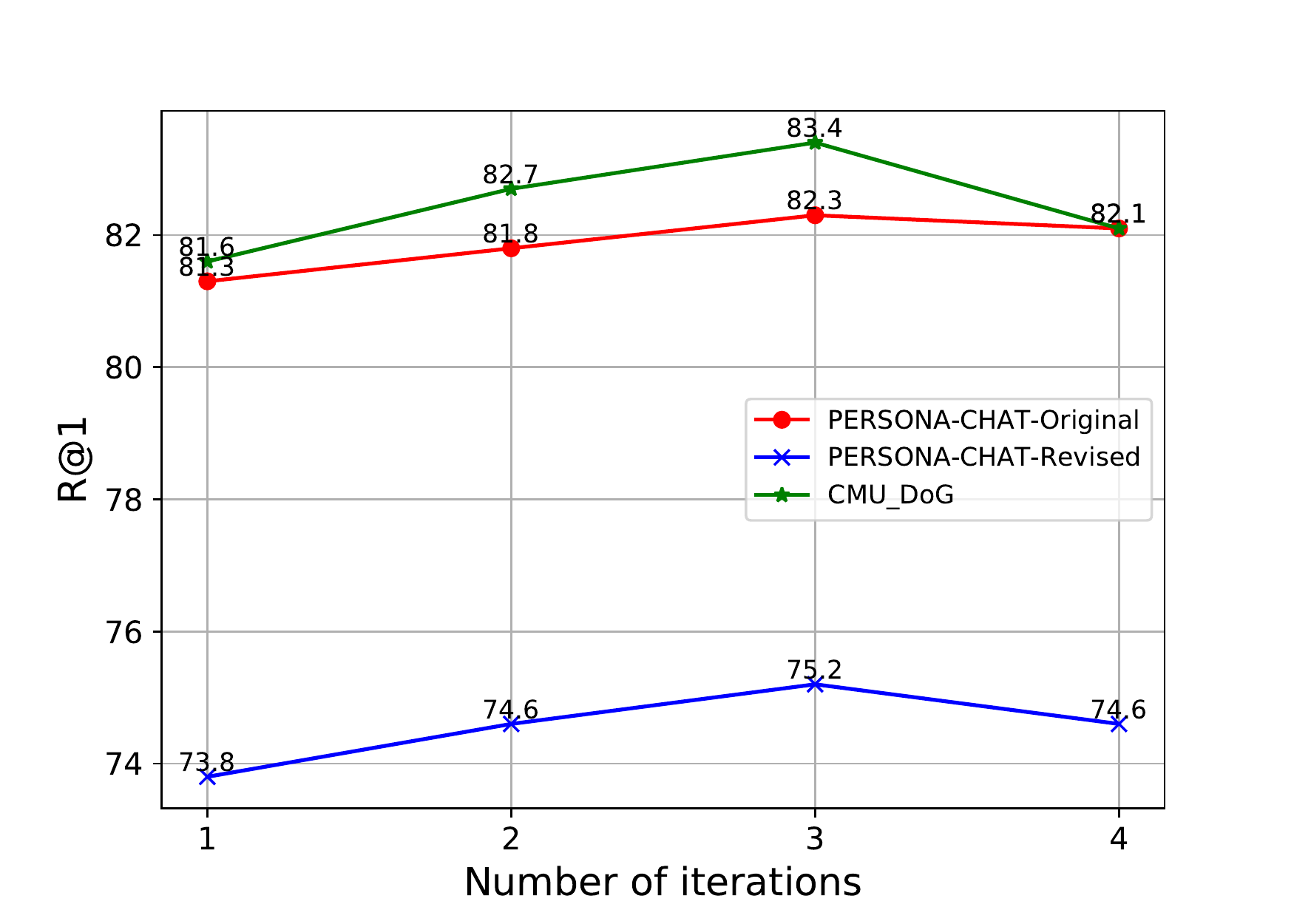}
        \caption{Validation set performance of FIRE  with  different number of iterations in iteratively referring.}
        \label{fig5}
      \end{figure}

      Figure~\ref{fig5} illustrates how the validation set performance of FIRE changed with respect to the number of iterations in iteratively referring.
      From it, we can see three iterations led to the best performance on both datasets.

    \paragraph{Complexity}

      We analysed the time complexity difference between FIRE and DIM.
      We recorded their inference time over the validation set of PERSONA-CHAT under the configuration of original personas using a GeForce GTX 1080 Ti GPU.
      It takes FIRE 109.5s and DIM 160.4s to finish the inference, which shows that FIRE is more time-efficient.
      The reason is that we design a lighter aggregation method in FIRE by replacing recurrent neural network in the aggregation part of DIM with a single-layer non-linear transformation.

\section{Conclusion}

  In this paper, we propose a method named \textbf{F}iltering before \textbf{I}teratively \textbf{RE}ferring (\textbf{FIRE}) for utilizing the background knowledge of dialogue agents in retrieval-based chatbots.
  In this method, a context filter and a knowledge filter are first designed to make the representations of context and knowledge aware of each other.
  Second, an iteratively referring network is built to collect deep and comprehensive matching information for scoring response candidates.
  Experimental results show that FIRE achieves a new state-of-the-art performance on two datasets.
  In the future, we will explore better ways of integrating pre-trained language models into our proposed methods for knowledge-grounded response selection.

\section*{Acknowledgments}
  We thank the anonymous reviewers for their valuable comments.
  This work was supported in part by the National Key R\&D Program of China under Grant 2019YFF0303001, and in part by the National Nature Science Foundation of China under Grant 61871358 and Grant U1636201.
  The last author's research is supported by NSERC Discovery Grants and Discovery Accelerator Supplement Grants (DAS).

\bibliography{emnlp2020}
\bibliographystyle{acl_natbib}

\end{document}